\documentclass[letterpaper, 10 pt, conference]{ieeeconf}
\usepackage{xcolor,soul}
\usepackage{graphicx} 
\usepackage{amsmath} 
\usepackage{amsfonts}
\usepackage{booktabs}
\usepackage{etoolbox}
\usepackage{multirow}
\usepackage{xparse}
\IEEEoverridecommandlockouts{}
\definecolor{ulsim}{HTML}{FDA736}
\definecolor{ulconverted}{HTML}{D35072}
\definecolor{ulboth}{HTML}{8606A7}
\definecolor{ulreal}{HTML}{0D0887}

\setul{1pt}{.9pt}
\DeclareRobustCommand{\ulsim}[1]{\setulcolor{ulsim}\ul{#1}}
\DeclareRobustCommand{\ulconverted}[1]{\setulcolor{ulconverted}\ul{#1}}

\DeclareRobustCommand{\ulreal}[1]{\setulcolor{ulreal}\ul{#1}}

\makeatletter
\patchcmd{\@makecaption}
  {\\}
  {.\ \ }
  {}
  {}
\makeatother

\newcommand{\argmax}{\operatornamewithlimits{argmax}}
\newcommand{\argmin}{\operatornamewithlimits{argmin}}
\newcommand{\dataindex}[1]{\text{\tiny{$#1$}}}
\newcommand{\ra}[1]{\renewcommand{\arraystretch}{#1}}
\NewDocumentCommand{\rot}{O{90} O{0.8em} m}{\makebox[#2][l]{\rotatebox{#1}{#3}}}%

\newcommand{\genesis}{GeneSIS-RT}

\title{\LARGE \bf
GeneSIS-RT: Generating Synthetic Images for training Secondary Real-world Tasks
}
\author{Gregory J. Stein and Nicholas Roy
  \thanks{Massachusetts Institute of Technology, 77 Massachusetts Ave, Cambridge, MA 02139 USA {\tt\small gjstein, nickroy@mit.edu}}%
\thanks{This work was supported by ARO under the Robotics Collaborative Technology Alliance and the Defense Advanced Research Project Agency (DARPA) under Contract No. HR0011-15-C-0110. Their support is gratefully acknowledged. G. J. Stein acknowledges support by a NDSEG Graduate Fellowship.}%
}

\begin{document}
\maketitle
\thispagestyle{empty}
\pagestyle{empty}

\begin{abstract}

We propose a novel approach for generating high-quality, synthetic data for domain-specific learning tasks, for which training data may not be readily available. We leverage recent progress in image-to-image translation to bridge the gap between simulated and real images, allowing us to generate realistic training data for real-world tasks using only unlabeled real-world images and a simulation. GeneSIS-RT ameliorates the burden of having to collect labeled real-world images and is a promising candidate for generating high-quality, domain-specific, synthetic data.

To show the effectiveness of using GeneSIS-RT to create training data, we study two tasks: semantic segmentation and reactive obstacle avoidance. We demonstrate that learning algorithms trained using data generated by GeneSIS-RT make high-accuracy predictions and outperform systems trained on raw simulated data alone, and as well or better than those trained on real data. Finally, we use our data to train a quadcopter to fly 60 meters at speeds up to 3.4 m/s through a cluttered environment, demonstrating that our GeneSIS-RT images can be used to learn to perform mission-critical tasks.

\end{abstract}

\section{Introduction}
\label{sec:intro}

As monocular cameras have become increasingly cheap and lightweight, they have become standard sensors for robotic systems and promise to enable capabilities necessary for high-level autonomy. Image data is rich with information and can be used for tasks ranging from object detection and semantic segmentation to reactive obstacle avoidance. Yet as the demand for domain-specific learning systems grows, so too does the demand for large volumes of labeled training image data, which such systems require if they are to learn to perform well. Publicly available datasets are limited in scope, so domain-specific applications typically require custom data that can be expensive to obtain.

\begin{figure}[t]
  \centering
  \includegraphics[width=\columnwidth]{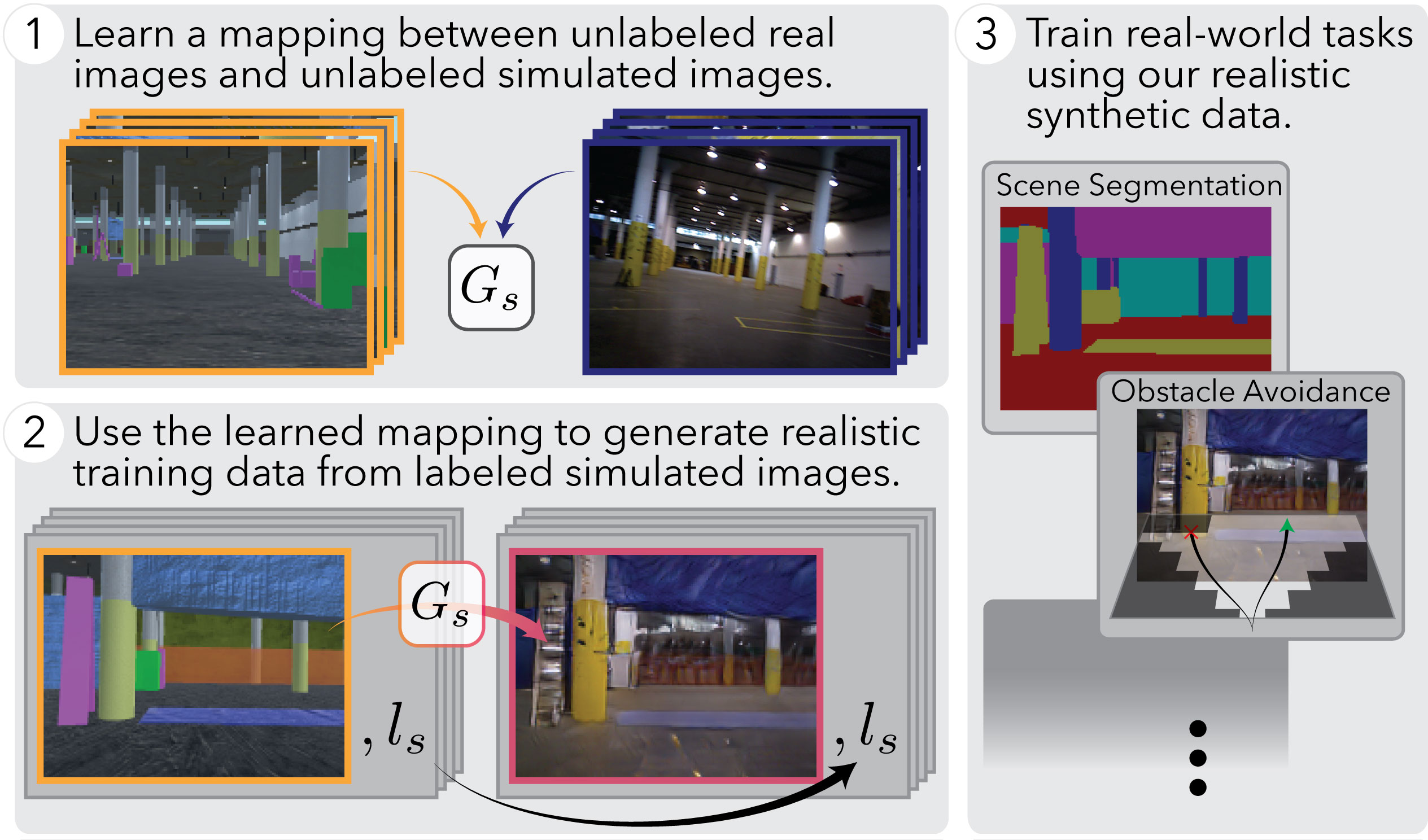}
  \caption{This schematic shows the procedure for \genesis{}. Upon generating labeled simulated images, we use a learned mapping function $G_s$ to make the images more realistic. Our converted images are suitable for training secondary real-world tasks, like semantic segmentation and reactive obstacle avoidance.}\label{fig:genesis-rt-flowchart}
\end{figure}

For example, tasks like semantic segmentation and reactive obstacle avoidance are inference problems of the form
\begin{equation}\label{eq:max-likelihood-label}
  l^* = \argmax_{l \in L} p(l | i ; D),
\end{equation}
in which we choose the maximum likelihood label $l^* \in L$ given an $m \times n \times d$ image $i \in \mathbb{R}^{m \times n \times d}$. A training dataset $D$, which consists of $N$ labeled images: $D = {\{ i^\dataindex{k}, l^\dataindex{k} \}}_{k=1}^N$, can be used used to learn the hyperparameters of the distribution (e.g., the weights of the neural network). The image label $l$ may be per-pixel, as in semantic segmentation, or per-image, as in choosing safe trajectories for reactive obstacle avoidance. At test time, the images $i$ are drawn from the set of real-world images: $i_r \sim \mathcal{I}_r$, where the $r$ subscript denotes \emph{real-world}. Ideally, we would \emph{train} with labeled real-world image data as well, $D_r = \{ i_r^\dataindex{k}, l_r^\dataindex{k} \}_{k=1}^N$, so as to match the test data. But labeling images to train a model for semantic segmentation or obstacle avoidance can be expensive, especially if the image labeling must be done by hand---hand-annotating a single image for training semantic segmentation can take up to 20 minutes, and thousands of labeled images may be required to learn the target concept. Simulation tools like the Unity~\cite{unity} or Unreal~\cite{unreal} game engines can alleviate the burden of data collection. Such tools can generate large volumes of simulated images $i_s \sim \mathcal{I}_s$, where the $s$ subscript denotes \emph{simulated}, from an underlying 3D model. Simulations use geometry and semantic concepts like objects when generating images, which makes it easy to simultaneously label individual pixels or the overall image for different tasks. But using simulated data for training assumes $p(l|i; D_s) \approx p(l | i; D_r)$, implying that the label we choose at test time would still have a high likelihood if we had trained with real data instead.

Training on simulated data and testing in the real world does not work in general, owing to the difficulties associated with producing photorealistic simulated images of arbitrary environments. Real and simulated images frequently differ in texture, lighting and color, features upon which modern learning algorithms, like convolutional neural networks, frequently rely. Suppose there existed a mapping function $G_s: \mathcal{I}_s \rightarrow \mathcal{I}_r$, which could convert a simulated image $i_s$ into its real-world counterpart: a \emph{converted} image $i_c = G_s(i_s)$. Assuming we had access to such a mapping, we could generate realistic training data from simulation $D_c = {\{ i_c^\dataindex{k}, l_s^\dataindex{k} \}}_{k=1}^N$, with which we can much more closely approximate $p(l | i; D_r)$.

In this work, we make use of recent progress in image-to-image translation to learn a mapping function $G_s$ and use it to generate more realistic synthetic training images. We show that the CycleGAN unpaired image-to-image translation procedure is a good candidate for learning $G_s$ and that can $G_s$ learned using \emph{unlabeled} real images---an advantage since collecting unlabeled images is easy compared to the process of labeling them. We can generate labeled ``converted'' images that are indistinguishable from labeled real images by generating labeled synthetic images from our simulation environment and then applying the mapping function $G_s$. We establish the effectiveness of using our \emph{converted} images on two tasks: semantic segmentation and reactive obstacle avoidance. For both tasks, we show that networks trained on our converted data outperform those trained on simulated data alone.

\section{Learning the Mapping Function}
\label{sec:choices-g_s}

The recent CycleGAN algorithm~\cite{CycleGAN2017} can learn the mapping function $G_s$ using \emph{unpaired} batches of images and can discover the relationship between two distributions of image types even when presented with novel images not in its training data. This mapping ability allows us to use CycleGAN to learn an effective mapping function $G_s$ provided only unpaired, unlabeled images from the simulator and real world---thus allowing us to produce realistic synthetic images with minimal overhead.

\subsection{The CycleGAN Procedure}

At the core of the CycleGAN procedure are two generative neural networks that can map simulated images to the set of real images and \emph{vice versa}. Each generative network is trained as part of a \emph{Generative Adversarial Network} (GAN)~\cite{goodfellow2014generative}, which involves creating a pair of networks and training them \emph{against one another} as part of an adversarial game. The first network in each $GAN$, the generator $G_s$, learns a mapping from $m \times n \times d$ images from set $\mathcal{I}_A \in \mathbb{R}^{m \times n \times d}$ and $m \times n \times d$ images from set $\mathcal{I}_B \in \mathbb{R}^{m \times n \times d}$ so that $G_s: \mathcal{I}_A \rightarrow \mathcal{I}_B$. The other network, the discriminator $D_s$, is structured as a classifier, which aims to tell the difference between images actually drawn from $\mathcal{I}_B$ and the images produced by $G_A$: i.e., $D_B: \mathcal{I}_B \rightarrow [0, 1]$. As such, the GAN loss is given by:
\begin{equation}\label{eq:gan-loss}
  \begin{array}{r@{}l}
  \mathcal{L}_{\text{GAN}}(G_A, D_B; &\mathcal{I}_A, \mathcal{I}_B) = \mathbb{E}_{i_B \sim \mathcal{I}_B}\left[ \log\left( D_B(i_B)\right) \right] + \\
  & \mathbb{E}_{i_A \sim \mathcal{I}_A}\left[ \log\left(1 -  D_B(G_A(i_A))\right) \right].
  \end{array}
\end{equation}
During each training iteration, the generator network is given a set of images from $\mathcal{I}_A$ and produces a corresponding set of converted images. These converted images and a set of images from $\mathcal{I}_B$ are given to the discriminator network for it to classify. The parameters of the two networks are tuned using backpropagation---the weights of $G_A$ are updated so as to produce more convincing converted images to fool $D_B$ (minimizing Eq.~\eqref{eq:gan-loss}) while the weights of $D_B$ are tuned to better differentiate between the two (maximizing Eq.~\eqref{eq:gan-loss}). As the system is trained, the converted images produced by $G_A$ increasingly resemble images from $\mathcal{I}_B$.

For our purposes, we would like to learn a generator $G_s$ capable of converting simulated images into realistic images. We also want this mapping function to preserve high-level structure: we do not want a large object, like a building, to vanish after conversion, since such objects are often important for determining the label $l$ we aim to predict. However, the loss function in Eq.~\eqref{eq:gan-loss} does not explicitly encourage this behavior. The CycleGAN procedure~\cite{CycleGAN2017} addresses this problem. First, it introduces a second pair of networks, $G_B: \mathcal{I}_B \rightarrow \mathcal{I}_A$ and $D_A: \mathcal{I}_A \rightarrow [0, 1]$, and a corresponding loss function, $\mathcal{L}_{\text{GAN}}(G_B, D_A; \mathcal{I}_B, \mathcal{I}_A)$, such that the networks are trained in parallel to the first, but for the reverse mapping. This second set of networks makes CycleGAN \emph{reversible}. A third loss term, the \emph{cycle-consistency loss}, is added as well: a per-pixel loss encouraging that the composition of the generator networks be the identity\footnote{By contrast, previous efforts~\cite{DBLP:journals/corr/ShrivastavaPTSW16} used a heuristic per-pixel loss after applying \emph{only one} of the generator functions. Their forward-only loss term tries to enforce high-level structure but at the expense of the quality of the converted images, an issue the CycleGAN procedure avoids.}:
\begin{equation}
  \label{eq:cycle-loss}
  \begin{array}{r@{}l}
    \mathcal{L}_{\text{cyc}}(G_A, G_B) = & \mathbb{E}_{i_A \sim \mathcal{I}_A}\left[\lVert i_A - G_B(G_A(i_A)) \rVert_1 \right] + \\
                                         & \mathbb{E}_{i_B \sim \mathcal{I}_B}\left[\lVert i_B - G_A(G_B(i_B)) \rVert_1 \right].
  \end{array}
\end{equation}
Our total loss is a sum of these three terms:
\begin{equation}
  \label{eq:cyclegan-loss}
  \begin{array}{r@{}l}
  \mathcal{L}(G_A, G_B, D_A, D_B) =& \lambda \mathcal{L}_{\text{cyc}}(G_A, G_B) + \\
  &\mathcal{L}_{\text{GAN}}(G_A, D_B; \mathcal{I}_A, \mathcal{I}_B) + \\
  &\mathcal{L}_{\text{GAN}}(G_B, D_A; \mathcal{I}_B, \mathcal{I}_A)
  \end{array}
\end{equation}
where $\lambda$ determines the relative importance of the GAN loss and the cycle-consistency loss. The loss is optimized such that
\begin{equation}
  \label{eq:1}
  G_A^*, G_B^* = \argmin_{G_A, G_B} \max_{D_A, D_B} \mathcal{L}(G_A, G_B, D_A, D_B)
\end{equation}

This procedure allows us to generate a mapping between unpaired, unlabeled sets of simulated and real images, so that we may then generate large batches of labeled realistic data from labeled synthetic data for training secondary real-world tasks. We empirically show that the resulting mapping function $G_s$ transfers more local features, like texture, while preserving macroscopic features of the scene, like objects, in Sec.~\ref{sec:cycl-conv-results}.

\subsection{Training $G_s$}
\label{sec:training-cyclegan}

We use the open-source implementation of the CycleGAN algorithm~\cite{CycleGAN2017} for our experiments. The \emph{cycle loss weighting}, which determines the magnitude of the \emph{cycle consistency loss}, was reduced from 10 to 4, so as to increase the relative importance of the discriminator network---this slightly reduced convergence time but made little difference in overall conversion quality. We also used the \emph{6-resent-block} configuration for the structure for the generative networks. We refer the reader to the work of Zhu \emph{et al}~\cite{CycleGAN2017} for further details on the network architecture and parameters.

\section{Generating \genesis{} Data}
\label{sec:real-simul-envir}

We should be able to learn a mapping function $G_s$ for a wide variety of environments.
In this work we focus on two different spaces, a warehouse environment and an outdoor environment, so that we may demonstrate the effectiveness of our technique. We emphasize that these spaces and the tasks for which they are used are merely a sample of the possible application domains of \genesis{}.

We rely on the Unity game engine to generate our simulated data $D_s$---while many of our simulated objects are unrealistic in both color and texture, the mapping function $G_s$ should learn the relationship between them and their real-world counterparts, thus eliminating the need for a user to spend days (or even weeks) modeling the appearance each object to appear photorealistic.

To learn the mapping function, we require unlabeled images from both the real and simulated environments. Collecting the real image data required hand-carrying an ASUS Xtion camera\footnote{The Xtion also comes with depth information, which we use in Sec.~\ref{sec:react-obst-avoid} for collecting the ``ground truth'' collision labels. For training $G_s$, however, only the RGB images are used.} around the different spaces, and roughly 40,000 images were collected for each. Simulated images were generated at random with a uniform distribution over our different environments. The parameters of the simulated camera were set to match those of the Xtion sensor---we generated on the order of 30,000 simulated images for each environment, all of which were used for training the mapping function $G_s$.

\subsection{The Warehouse Environment}
\label{sec:wareh-envir}

\begin{figure}[t]
  \centering
 \includegraphics[width=0.90\columnwidth]{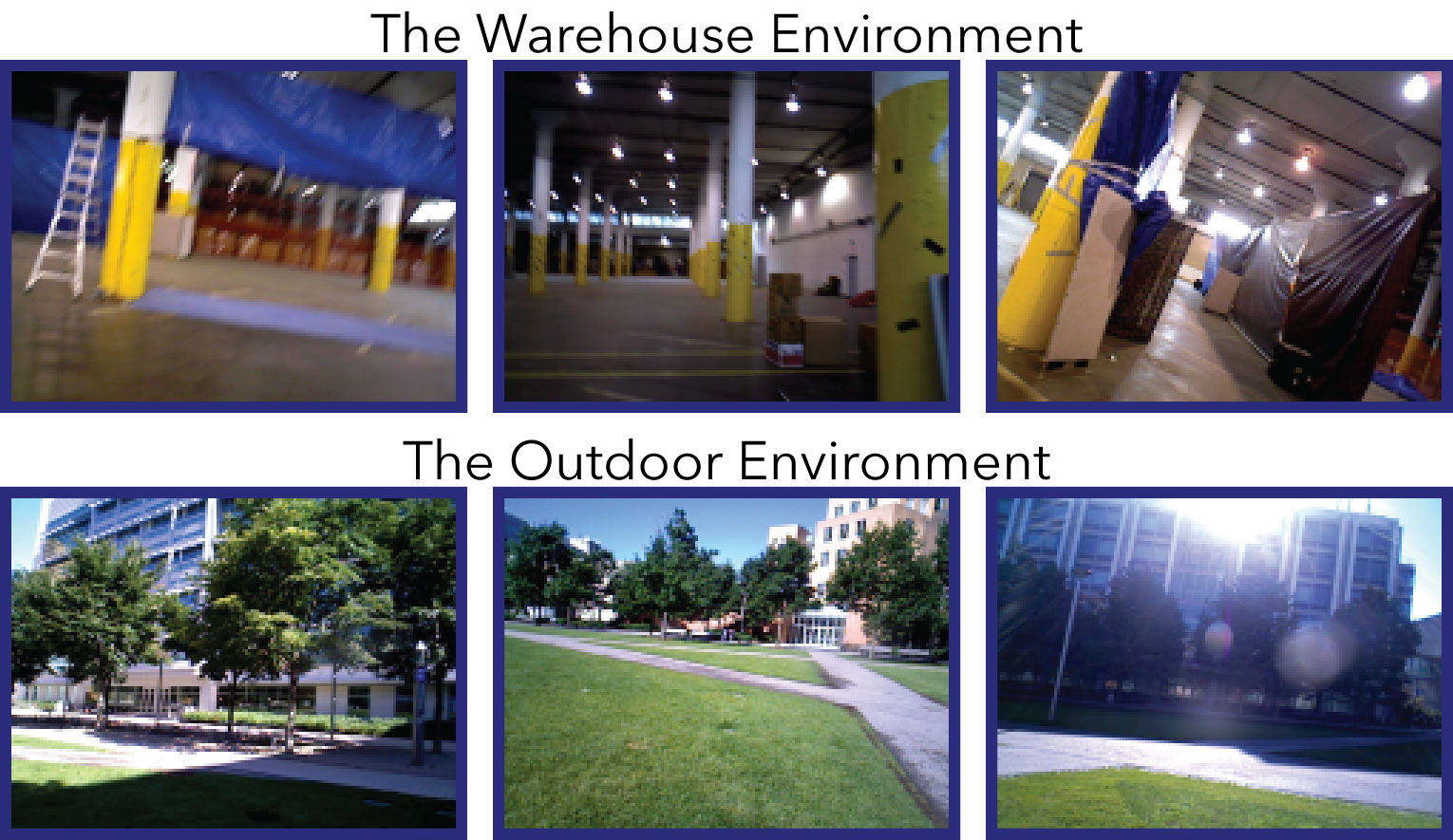}
 \caption{Here we show a few \ulreal{real-world images} from each of the two environments we use for testing. More details about these environments can be found in Sec.~\ref{sec:wareh-envir} and Sec.~\ref{sec:outdoor-environment}.}\label{fig:real-environments}
\end{figure}

The first environment is a warehouse---images from the real warehouse can be found on the top in Fig.~\ref{fig:environments-and-conversion-results}. Besides a narrow tarp-constructed hallway, the space consists mostly of a lattice of large pillars. The warehouse also includes some relatively sparse clutter, like divider panels, cardboard and plastic boxes, and additional tarps. In the simulation, regions of the environment outside the center area of interest, which are perceptible by the camera yet otherwise unimportant, are replaced with textured walls, thereby eliminating the need for the user to have to model a larger and more-complicated space. It is expected that the mapping function will learn the relationship between these synthetic boundaries and their corresponding regions in the real world.

\subsection{The Outdoor Environment}
\label{sec:outdoor-environment}

Our other test environment is the ``North Court'' courtyard on the MIT campus---images from the real outdoor environment can be found on the bottom in Fig.~\ref{fig:environments-and-conversion-results}. The courtyard consists of a grassy area bordered by trees and surrounded by a few architecturally diverse buildings. Our simulated environment contains CAD models of the campus buildings from the MIT facilities department placed on overhead satellite imagery of MIT~\cite{TheNationalMap}. All other objects were modeled by the authors using simple textures and 3D primitives.

\subsection{Image Conversion Results}
\label{sec:cycl-conv-results}

\begin{figure}[t]
  \centering
  \includegraphics[width=1.00\columnwidth]{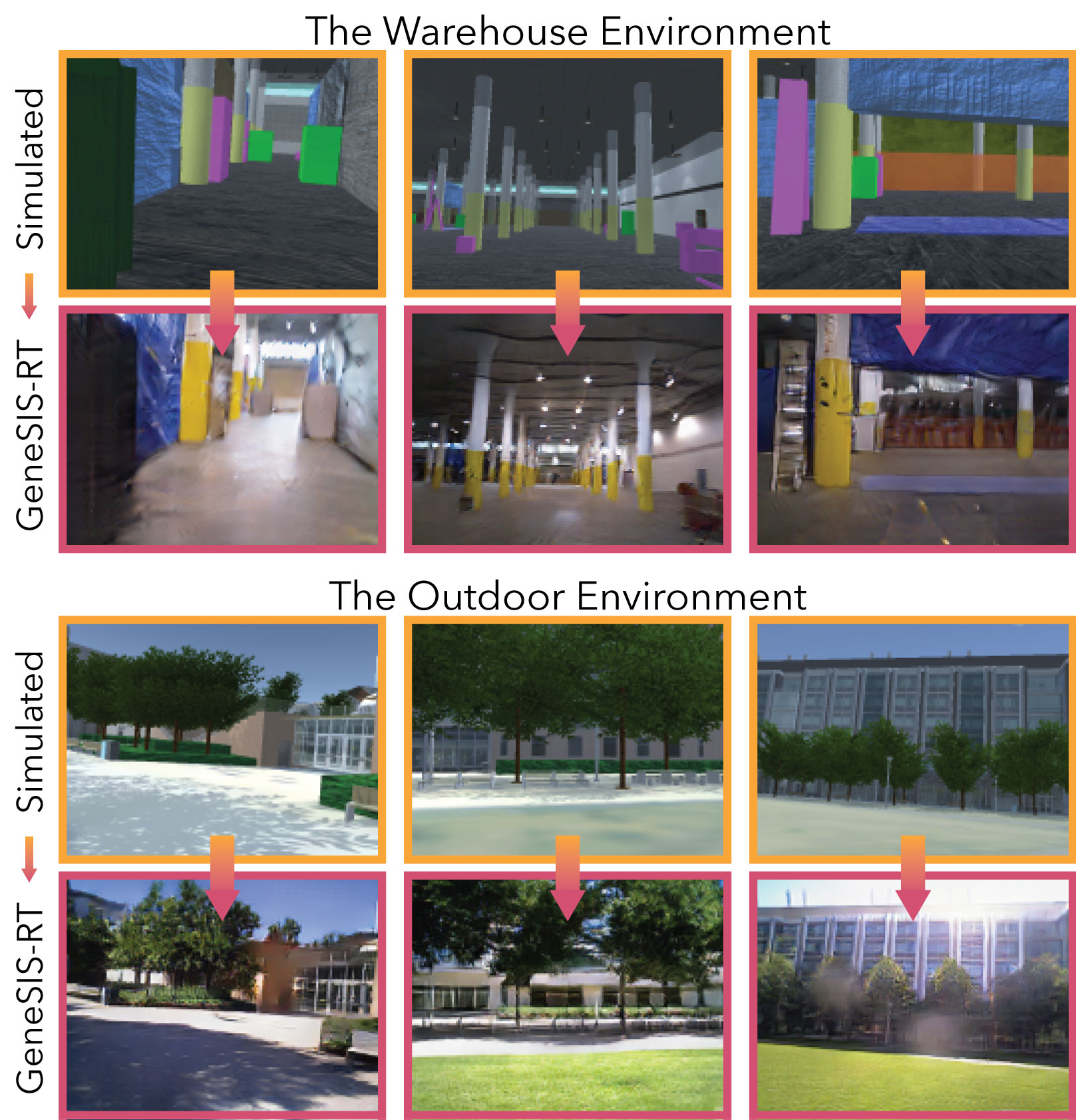}
 \caption{Here we show \ulsim{simulated images} alongside their \ulconverted{GeneSIS-RT image counterparts}, obtained by applying our learned mapping function $G_s$, for each of the two environments we use for testing. Our \genesis{} process yields images that look much more realistic and that are better suited for training learning algorithms for real-world tasks. In the \ulconverted{converted images} we observe diffuse reflections on the warehouse floor and dramatically improved shadows outdoors, realistic features not present in the \ulsim{simulated images}. We later evaluate the quality of our \genesis{} data in the context of specific tasks, focusing on semantic segmentation in Sec.~\ref{sec:semant-segm} and reactive obstacle avoidance in Sec.~\ref{sec:react-obst-avoid}.}\label{fig:environments-and-conversion-results}
\end{figure}

Converted images from both the warehouse and outdoor environments can be found alongside their simulated image counterparts in Fig.~\ref{fig:environments-and-conversion-results}. Though it is difficult to quantitatively evaluate the \emph{realism} of an image, a qualitative inspection of the converted images shows that they have real-world features not present in the simulated images. The conversion process correctly matches textures, like those of pillars and buildings, and also picks up on more subtle features, like the diffuse reflection of sunlight on the smooth warehouse floor. In the warehouse environment, the network also correctly associates the simulated green obstacles with the tall, gray dividers and the pink objects with black and cardboard boxes. The results for the outdoor environment are similar---beyond simply improving color, the algorithm makes the structure of the converted trees more closely match that of their real-world counterparts and enhances the realism of shadows.

While the images look more realistic, minor oddities can be observed for both environments. First, less common objects, like the ladders in the warehouse, sometimes convert poorly. Though they resemble their physical counterparts, they can occasionally blend in with the scene behind them. Second, the conversion process can fail whenever any one object takes up a large portion of the image, since the mapping algorithm aims to understand a more complex scene. When simulated images containing only a single tarp or obstacle are passed to $G_s$, the resulting images often contain artifacts, like small bursts of color. However, neither effect is sufficiently common so as to hurt overall performance of our secondary learning algorithms in practice.

Ultimately, the best metric for how realistic our images look, i.e., how closely they resemble real images drawn from $\mathcal{I}_r$, is how useful they are as training data for secondary tasks. As such, the next two sections are devoted to the evaluation of networks trained on our converted data for semantic segmentation and reactive obstacle avoidance.

\section{Application to Semantic Segmentation}
\label{sec:semant-segm}

\begin{figure}[t]
  \centering
 \includegraphics[width=\columnwidth]{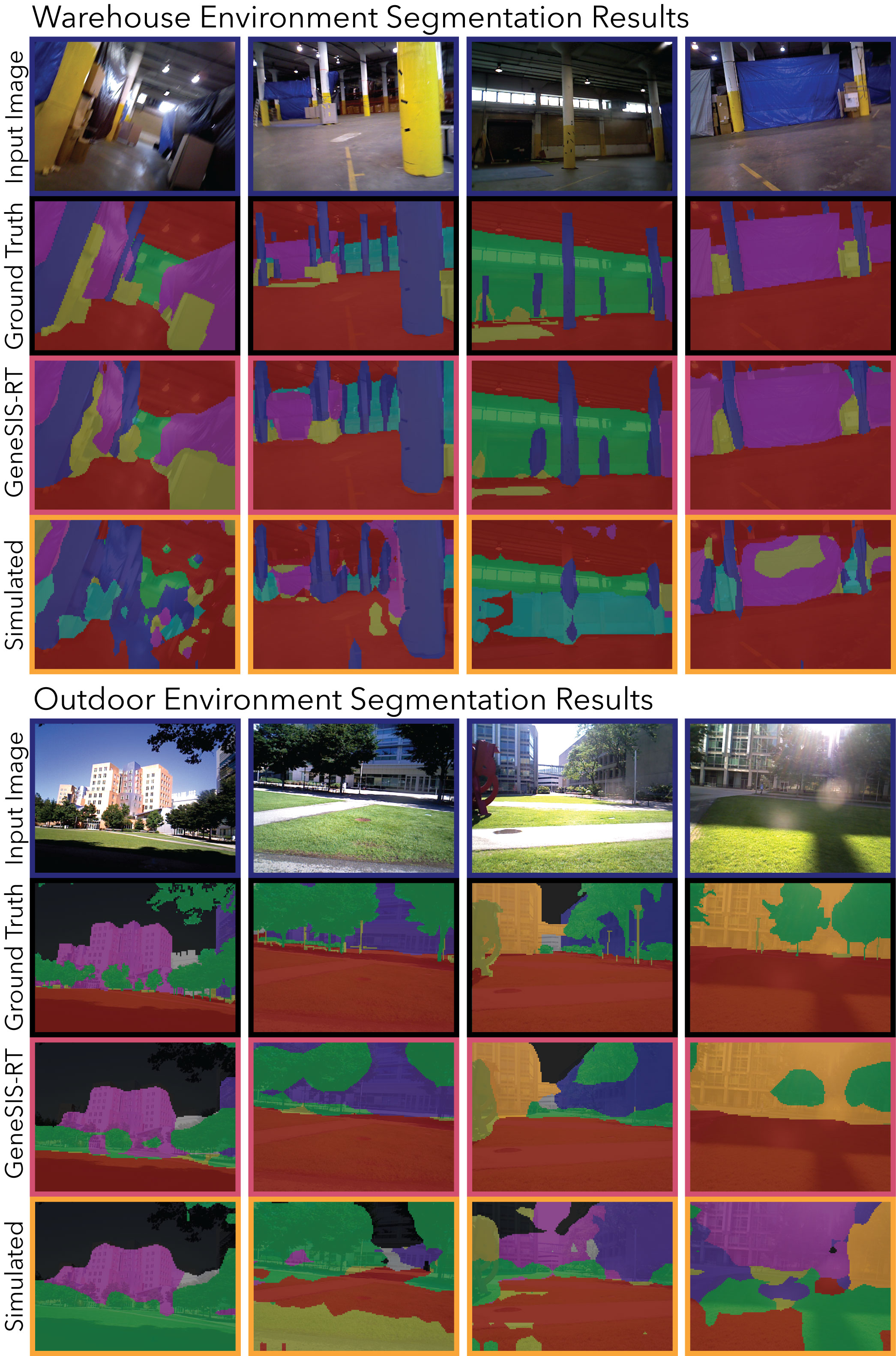}
 \caption{Here we see the results of training the semantic segmentation network on our datasets for each of the warehouse and outdoor environments. Each column of four images makes up a set comprised of the input image, the ground truth segmentation (hand-annotated by the authors), the prediction from a network trained on \ulsim{simulated data}, and the prediction from a network trained on our \ulconverted{GeneSIS-RT data}. It is clear to see in all cases that the converted images are far more useful for training in these environments than the simulated images. The \ulconverted{GeneSIS-RT network} predictions are far more accurate at correctly labeling the buildings in the outdoor environment, which each have a unique class. Details about the segmentation task can be found in Sec.~\ref{sec:semant-segm}.}\label{fig:segmentation-results}
\end{figure}

Once we have generated the converted images $i_c$, they can be used to train separate neural networks to perform specific tasks. First, we focus on \emph{semantic segmentation}, in which each pixel of an $m \times n$ input image is assigned a classification label (e.g., floor, wall, tree), so that $l \in \mathcal{C}^{m \times n}$ where $\mathcal{C} = \{c_0, c_1, \cdots, c_n\}$ are the possible classes. Semantic segmentation is well suited for using \genesis{} data, since hand-annotating real-world images is expensive and time consuming and the simulator can provide the class ID for each pixel in the converted images. Semantic segmentation also relies heavily on accurate color, lighting, and texture, features that are a challenge to produce directly from simulation but which $G_s$ provides. Furthermore, though there are public training datasets for this task, existing datasets of real images typically provide only a fixed set of label classes, which restricts their use for tasks requiring a higher degree of specificity.

\subsection{Data Generation \& Network Training}
\label{sec:gener-segm-data}

We create two datasets for testing: one with images directly from the simulated environment $i_s$ and one with the \genesis{} converted images $i_c = G_s(i_s)$. The datasets each consist of roughly 20,000 image/label pairs at a resolution of $160 \times 120$ pixels. The segmentation data was derived from our Unity simulation environments, which makes it easy to generate pixel-accurate labels associated with each observation.

For each environment, we used a set of class labels that are unique to our environments, which means that no off-the-shelf dataset exists for our use case. Obtaining the necessary data without the assistance of a simulation tool would require a substantial effort in hand-annotating data. For the warehouse environment, we used 6 label classes, including tarps (magenta) and pillars (blue). For the outdoor environment, we assigned a unique class label to each building, in addition to standard classes like trees and clutter.

For our neural networks, we used DeepLabv2, a variant of ResNet-101 usable for semantic segmentation~\cite{CP2016Deeplab,TensorflowDeepLab,tensorflow2015-whitepaper}. The network was pre-trained on the MS-COCO dataset~\cite{502COCO}, and we used these weights as a starting point for our training. We trained using the provided script, with a batch size of 10, 20,000 iterations, and an initial learning rate of $2.5\times 10^{-4}$. For more details, we refer the reader to the implementation~\cite{TensorflowDeepLab}.

\subsection{Segmentation Results}
\label{sec:segmentation-results}
\setlength{\tabcolsep}{4pt}
\begin{table}
  \vspace*{4pt}
  \centering
  \caption{Outdoor Segmentation Results [mIoU]}\label{tab:outdoor-segmentation-results}
  \vspace*{-6pt}
    \begin{tabular}{@{}rrrrrrrrrrr|rr@{}}\toprule
      \rot{\sc{}Class Label} & \rot{Sky} & \rot{Ground} & \rot{Foliage} & \rot{Clutter} & \rot{X6 Buildings} & \rot{Stata Center} & \rot{Building 54} & \rot{Building 76} & \rot{Building 68} & \rot{Other Buildings} & \rot{Class Average} & \rot{Global Average} \\\midrule
     {\sc{}\genesis{}} & {\bf{}65} &  {\bf{}88} & {\bf{}57} & {\bf{}27} & {\bf{}61} & {\bf{}50} & {\bf{}65} & {\bf{}62} & {\bf{}55} & {\bf{}33} & {\bf{}56} & {\bf{}67}\\
     {\sc{}Simulated} & 37 & 55 & 37 & 12 & 7 & 24 & 10 & 2 & 4 & 6 & 19 & 33\\
    \end{tabular}
\end{table}
\begin{table}
  \vspace*{4pt}
  \centering
  \caption{Warehouse Segmentation Results [mIoU]}\label{tab:warehouse-segmentation-results}
  \vspace*{-6pt}
    \begin{tabular}{@{}rrrrrrr|rr@{}}\toprule
      \rot{\sc{}Class Label} & \rot{Floor/Ceiling} & \rot{Walls/Windows} & \rot{Clutter} & \rot{Pillars} & \rot{Tarps} & \rot{Outside Area} & \rot{Class Average} & \rot{Global Average} \\\midrule
     {\sc{}\genesis{}} & {\bf{}86} & {\bf{}79} & {\bf{}46} & {\bf{}75} & {\bf{}78} & {\bf{}61} & {\bf{}71} & {\bf{}77}\\
     {\sc{}Simulated} & 73 & 62 & 4 & 51 & 42 & 23 & 43 & 53 \\
    \end{tabular}
\end{table}

The results of evaluating these two networks on real-world data can be found in Fig.~\ref{fig:segmentation-results}, which shows the network predictions alongside the ground truth labels, which were hand-annotated by the authors. The network trained on our \genesis{} data vastly outperforms the network trained on the data obtained directly from the Unity simulator: for the outdoor environment, the global mean Intersection over Union (mIoU) on real-world test data was 66.9 for our \genesis{} data and only 32.6 when the network was trained on raw simulated data (a perfect image segmentation has a mIoU of 100). In the warehouse, the \genesis{}-trained network outperformed the simulated network 77.4 to 52.7. The per-class mIoU values on the real-world test data for the segmentation task can be found in Table~\ref{tab:outdoor-segmentation-results} and Table~\ref{tab:warehouse-segmentation-results}.

Since it lacks the color and texture of the real world, the raw simulated data was ineffective, and the network trained with it partitioned the space in odd places and frequently labeled these regions incorrectly. While it was sometimes able to correctly identify the ground plane or larger objects such as tarps or tree canopies, a large portion of the predictions did not agree with the true label.

By contrast, the network trained on \genesis{} data was much more accurate. In the outdoor environment, this network was able to correctly identify the class labels of the different buildings with high accuracy, a task which baffled the network trained on raw simulated data. Furthermore, this network was also capable of finding small regions of clutter in the warehouse environment, including distant boxes and mats on the ground. These results show both the effectiveness and usefulness of our technique---without providing any real labeled images, we have trained a system that can perform well in the real world.

\section{Application to Reactive Obstacle Avoidance}
\label{sec:react-obst-avoid}

\begin{figure}[t]
 \centering
 \includegraphics[width=0.95\columnwidth]{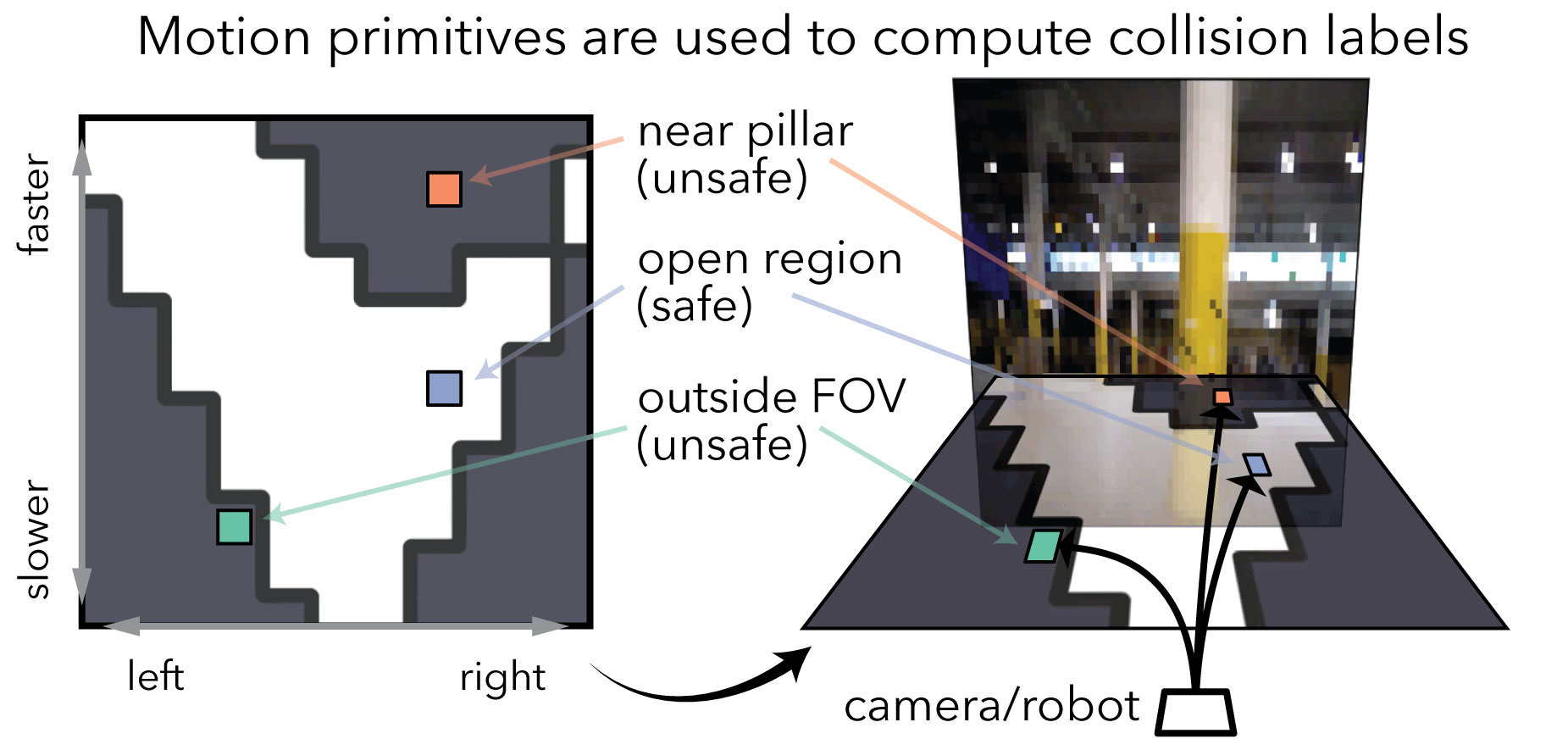}
 \caption{For the task of reactive obstacle avoidance, our set of actions corresponds to an $11 \times 11$ grid of target accelerations; the vehicle can perform some combination of speeding up, slowing down or banking left/right. Our neural network predicts a collision label for each potential action. We show the relationship between the motion primitives---which are used in combination with the 3D structure to compute the collision labels---and the camera image in the rightmost illustration above. For more details, see Sec.~\ref{sec:react-obst-avoid}.}\label{fig:motion-primitive-example}
\end{figure}

We also focused on the task of reactive obstacle avoidance, in which an aerial vehicle must use an image to predict which of a set finite-time actions might cause the vehicle to collide with the environment. In addition to the environment structure, which can be inferred from the image, the collision labels also depend on the vehicle's instantaneous velocity and acceleration. Incorporating the vehicle state requires only trivial modifications to Eq.~\eqref{eq:max-likelihood-label} and our learning problem therefore becomes:
\begin{equation}
  \label{eq:likelihood-obstacle-avoidance}
  l^* = \argmax_l p(l | i, v_0, a_0; D),
\end{equation}
where $v_0 \in \mathbb{R}^3$ is the vehicle's instantaneous velocity and $a_0 \in \mathbb{R}^3$ is its instantaneous acceleration, estimates of which are provided via an onboard state estimator~\cite{steiner2017samwise}. The label for the obstacle avoidance task is that of a multilabel binary classification problem with $N_c$ elements, so that $l \in {\{ 0, 1 \}}^{N_c}$ and the $n$th element of the label vector corresponds to the collision state of the $n$th action.

For our experiments, each action is a motion primitive specified by an acceleration target and defined by the dynamics model from~\cite{florence2016integrated}. Our set of motion primitives corresponds to an $11 \times 11$ grid of accelerations, allowing the vehicle to speed up, slow down or bank left/right to varying degrees, with a maximum acceleration of 2 m/s$^2$ along each axis. We show how the collision labels for these motion primitives and the geometry are related to one another in Fig.~\ref{fig:motion-primitive-example}.

The collision labels are computed from instantaneous depth data\footnote{We use the ASUS Xtion~\cite{xtion} to get depth data in the Warehouse environment and the Intel RealSense r200~\cite{realsense} to get depth data in the Outdoor environment.} using the procedure found in~\cite{florence2016integrated}, which simulates the trajectory of each 1.5 second motion primitive and checks for collision in 3D space. The label calculation also takes state estimator uncertainty into account, and we assign a collision label to ``near-misses'' as well. A label of 1 for any primitive indicates that the vehicle would pass too close to an obstacle or enter space that is out of view, either because it is occluded by an obstacle or because the action would take the vehicle outside the camera frustrum, and might therefore collide with an unseen obstacle. For further implementation details, we refer the reader to~\cite{florence2016integrated}.

The neural networks we use begin with two convolution/max-pool layers, which operate on color images scaled to $64 \times 48$ pixels. We then add a few fully-connected hidden layers---three for the outdoor environment and four for the warehouse---each using a ReLU activation. Finally, we include one final fully-connected layer with a sigmoid activation, and use a logarithmic loss function for training the system. All weight terms have an L2 loss, without which all three networks perform less well. The network architecture and regularization terms could likely be tuned further for improved performance, but exhaustively tuning these parameters is not the focus of this work.

\subsection{Obstacle Avoidance Performance}
\label{sec:obst-avoid-perf}

\ra{1.1}
\setlength{\tabcolsep}{4pt}
\begin{table}
  \vspace*{6pt}
  {\centering
  \caption{Results from the Obstacle Avoidance Task}\label{tab:obstacle-avoidance-metrics}
  \vspace*{-6pt}
{\begin{tabular}{@{}lrrr@{}}\toprule
 {\sc{}Dataset} & Network & AUROC & Log Loss \\\toprule
 {\sc{}Warehouse Test 1}& \genesis{} (ours) & 0.945 & 0.300 \\
 \emph{Test data collected on the}& Simulated & 0.892 & 0.537 \\
 \emph{same day as training data}& Real & {\bf{}0.954} & {\bf{}0.279} \\\midrule
  \parbox[t]{3cm}{\multirow{3}{*}{{\sc{}Warehouse Test 2}}}
 & \genesis{} (ours) & {\bf{}0.942} & {\bf{}0.348} \\
 & Simulated & 0.864 & 0.580 \\
 & Real & 0.908 & 0.520 \\\midrule
  \parbox[t]{3cm}{\multirow{3}{*}{{\sc{}Warehouse Flights}}}
 & \genesis{} (ours) & {\bf{}0.970} & {\bf{}0.222} \\
 & Simulated & 0.932 & 0.534 \\
 & Real & 0.958 & 0.342 \\\toprule
  \parbox[t]{3cm}{\multirow{3}{*}{{\sc{}Outdoor Test}}}
 & \genesis{} (ours) & {\bf{}0.917} & {\bf{}0.407} \\
 & Simulated & 0.892 & 0.743 \\
 & Real & 0.915 & 0.422 \\\bottomrule
 \end{tabular}}\\[8pt]}
Here we show the results from the reactive obstacle avoidance task, which includes the area under the ROC curve (AUROC) and the log loss for networks trained with three sets of data: \ulconverted{our GeneSIS-RT data}, \ulsim{raw simulation data}, and \ulreal{real-world data}. Except for \emph{Warehouse Test 1}, for which the evaluation data was collected on the same day as the real-world training data, the network trained with our \genesis{} data outperforms the other two networks. These results are discussed in detail in Sec.~\ref{sec:react-obst-avoid}.
\end{table}

To evaluate the effectiveness of our technique, we train three different networks $\mathcal{N}$ for each environment:
\begin{enumerate}
\item \ulconverted{$\mathcal{N}[\text{GeneSIS-RT}]$}: Trained with converted images $i_c = G_s(i_s)$ paired with simulated labels;\@
\item \ulsim{$\mathcal{N}[\text{Simulation}]$}: Trained with simulated data;\@
\item \ulreal{$\mathcal{N}[\text{Real}]$}: Trained with labeled real images $i_r$, collected with RGB-D sensors.
\end{enumerate}
All three data sets undergo a data augmentation procedure, in which we randomly vary the instantaneous vehicle velocity during the label calculation. This is to ensure that the network properly learns the relationship between the velocity and the collision probabilities.

\subsubsection{Warehouse Obstacle Avoidance}
\label{sec:wareh-obst-avoid}

\begin{figure}[t]
  \centering
  \includegraphics[width=\columnwidth]{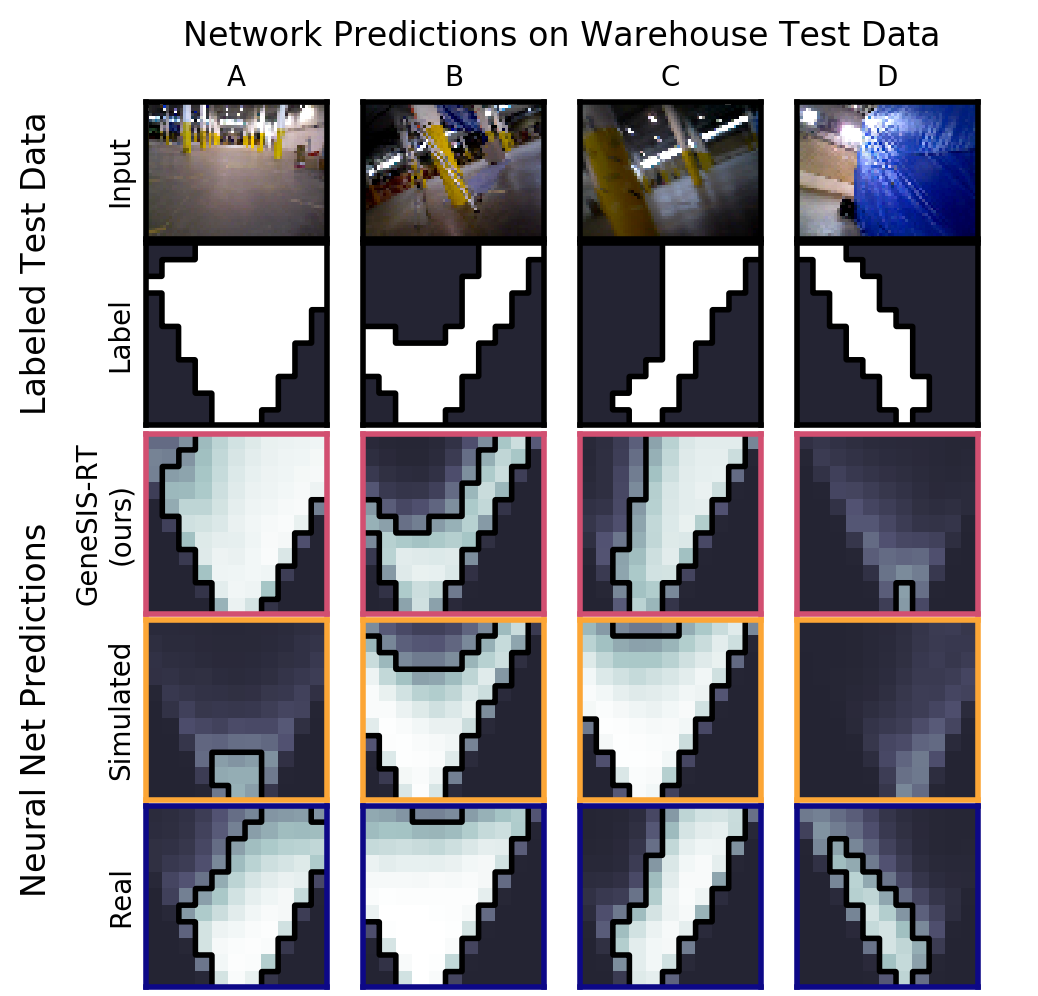}
  \caption{Here we show labeled image data for the task of reactive obstacle avoidance alongside predictions from neural networks trained using \ulconverted{our converted image data}, \ulsim{raw simulated images}, and \ulreal{real images}. The networks predict the probability of collision for each action: white denotes safe, charcoal denotes unsafe, and the intermediate colors represent degrees of confidence in the prediction (we also include a black contour at 50\% confidence). Our converted images much more closely resemble real-world images---not only are they clearly more useful for training than the simulated images, but the network trained using the \genesis{} data rivals, and even exceeds, the performance of its counterpart trained using real data. These results are discussed further in Sec.~\ref{sec:obst-avoid-perf}.}\label{fig:obs-avoid-warehouse-results}
\end{figure}

In the warehouse, we study the performance of these networks on three sets of evaluation data: two of these sets were collected by hand, emulating quadcopter flight, and the third, to be discussed in more detail in Sec.~\ref{sec:flight-performance} was collected during autonomous vehicle flight. The performance of the three networks on these datasets can be found in Table~\ref{tab:obstacle-avoidance-metrics}. Relatedly, we show a few individual network predictions alongside labeled images from the evaluation data in Fig.~\ref{fig:obs-avoid-warehouse-results}.

The \genesis{} network outperforms the $\mathcal{N}[\text{Simulated}]$ network. Though the raw simulated data matches the \genesis{} data in number of images and coverage of the warehouse space, the raw simulated images are not similar enough to real images to reliably make accurate predictions at test time. The $\mathcal{N}[\text{Simulated}]$ network frequently makes errors that would result in collision. By contrast, the \genesis{} data more closely resembles the real world and therefore enables more accurate predictions. The data in Table~\ref{tab:obstacle-avoidance-metrics} supports this conclusion: the network trained with \genesis{} data has reliably better cross entropy loss and area under the ROC curve than its counterpart trained on raw simulated data.

Furthermore, the network trained on \genesis{} data performs as well as and even exceeds the performance of the network trained on real data. The $\mathcal{N}[\text{Real}]$ network slightly outperforms the $\mathcal{N}[\text{GeneSIS-RT}]$ network on the \emph{Warehouse Test 1} dataset, which was collected on the same day and under the same conditions as the warehouse training data. The $\mathcal{N}[\text{GeneSIS-RT}]$ network demonstrates better generalization performance across the remaining two warehouse datasets, each collected on different days and under slightly different lighting conditions. The $\mathcal{N}[\text{GeneSIS-RT}]$ network outperforms the $\mathcal{N}[\text{Real}]$ network on the \emph{Warehouse Test 2} dataset and on the \emph{Warehouse Flight Test} dataset, which will discussed in more detail in Sec.~\ref{sec:flight-performance}.

We should note that none of the networks perform particularly well in the ``tarp hallway'' region of the warehouse. In this region, it is more difficult to determine precise distance to nearby obstacles, which are large and mostly uniform in color. We can see this effect in Fig.~\ref{fig:obs-avoid-warehouse-results}D---while both the $\mathcal{N}[\text{GeneSIS-RT}]$ and $\mathcal{N}[\text{Real}]$ networks direct the vehicle away from the observed obstacles, the confidence in the predictions is low, observed as the gray-blue color in the plots. This is a difficulty associated with the task of obstacle avoidance and should not be seen as a limiting factor in our overall approach.

\subsubsection{Outdoor Obstacle Avoidance}
\label{sec:outd-obst-avoid}

For the outdoor environment, we collected a single evaluation dataset by hand for testing, the results of which can be found at the bottom of Table~\ref{tab:obstacle-avoidance-metrics}. We can see that the performance of the \genesis{} trained network outperforms that of the network trained with raw simulated data, for both the AUROC and the log loss, and matches the quality achieved by real data as well. However, performance in this environment is not as good as it was in the warehouse for either network trained on our \genesis{} or real-world data, suggesting that we require higher-precision data to exceed this performance. However, our approach to generating data gives us access to a differentiable cost function over the space of images, something which is not possible with real-world data. Such a cost function may enable better performance, and is a subject for future research. Regardless, the results using our \genesis{} data are encouraging, despite the difficulty of the task.

\subsection{Real-World Quadcopter Flights}
\label{sec:flight-performance}

\begin{figure}[t]
  \centering
  \includegraphics[width=\columnwidth]{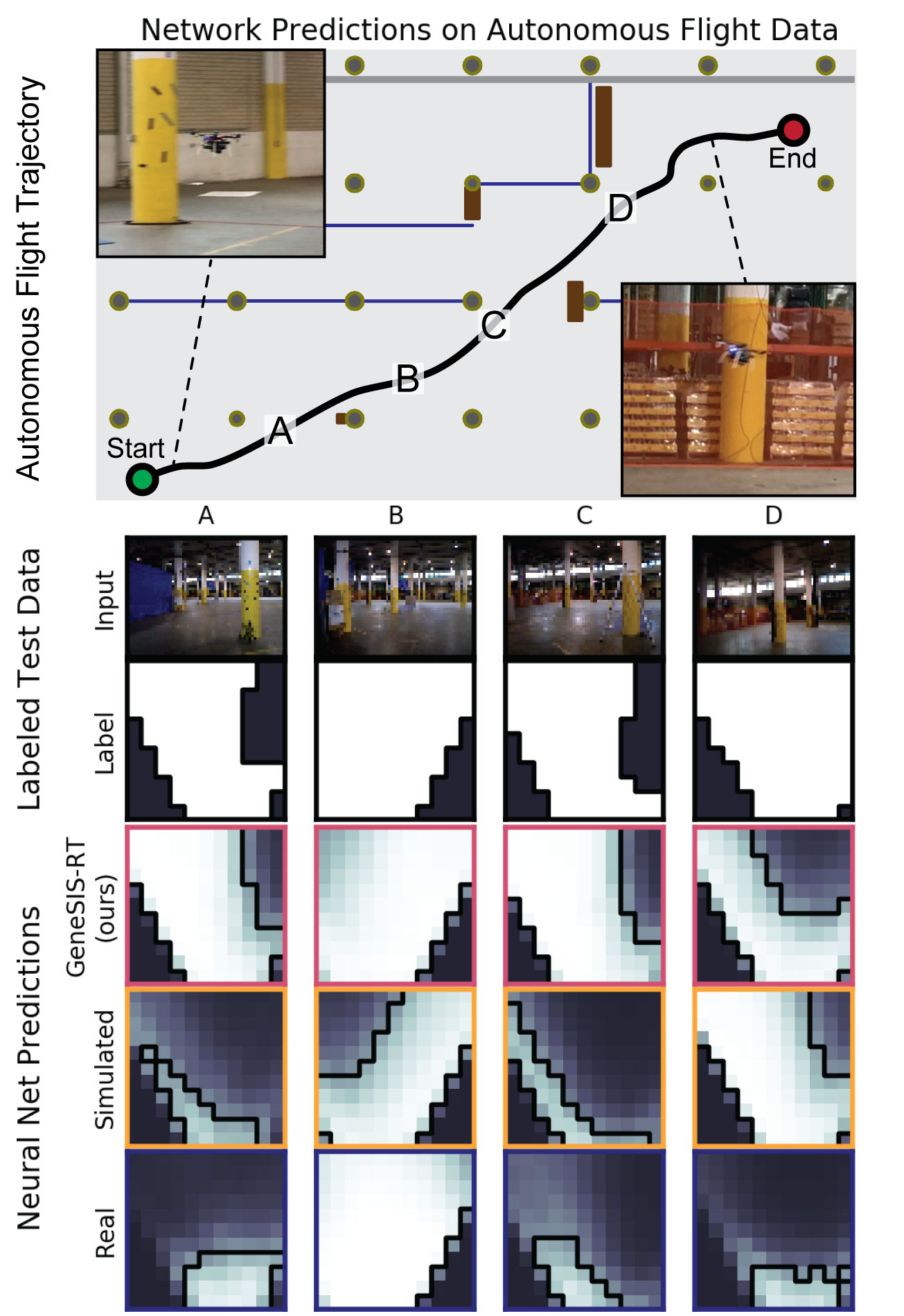}
  \caption{Here we show network predictions on data from an autonomous quadcopter flight, during which our \ulconverted{GeneSIS-RT data} was used for guidance and reactive obstacle avoidance. During the flight, the vehicle dodges five distinct obstacle, two of which can be seen in columns A and C, and reaches a top speed of 3.4 meters-per-second. The performance of the network trained with the \ulconverted{GeneSIS-RT data} clearly outperforms those trained on \ulsim{simulated data} and on \ulreal{real data}. Only once, shown in column D, does the vehicle dramatically slow down due to misclassified free space, and otherwise quickly and safely reaches the goal.}\label{fig:obs-avoid-warehouse-flight-results}
\end{figure}

To further test how well our converted data can be used to inform real-world decisions, we put our \genesis{} network on an autonomous robotic platform. We flew a physical quadcopter across our warehouse environment with the network in-the-loop, requiring that it dodge obstacles along the way. Success in this experiment relied on two things: (1) recognizing obstacles, so that it would not crash, and (2) reliably detecting free space, so that the vehicle could still make progress towards the goal.

We threshold the output predictions such that we only choose an action when the sigmoid output for a particular motion primitive is less than 0.05: i.e., we err on the side of slowing and stopping unnecessarily rather than risk collision. Thus the vehicle is expected to behave conservatively---taking wider trajectories around potential obstacles and slowing down or stopping as it becomes less certain about its safety. Furthermore, for safety reasons, the quadcopter is limited to speeds of no more than 3.5 $m/s$, yet is otherwise guided entirely by the decisions of the learned model.

We conducted 15 flights, covering over 250 meters. The vehicle successfully dodged dozens of obstacles, including pillars, tarps and clutter, and never crashed, all while correctly identifying free space with high enough accuracy to make forward progress. A comparison between the different networks on the flight data, shown in Table~\ref{tab:obstacle-avoidance-metrics} under \emph{Warehouse Flight Tests}, further demonstrates the effectiveness of using our technique---the network trained on our converted data outperforms the networks trained on raw simulated data and on real data.

The longest flight consisted of a single 60 meter path across the warehouse, during which the vehicle clearly avoided 4 obstacles and reached a top speed of 3.4 m/s. The data from this flight, consisting of network predictions from the $\mathcal{N}[\text{GeneSIS-RT}]$ network, which was guiding the vehicle, the $\mathcal{N}[\text{Real}]$ network, and $\mathcal{N}[\text{Simulated}]$ networks, is shown in Fig.~\ref{fig:obs-avoid-warehouse-flight-results}. The $\mathcal{N}[\text{GeneSIS-RT}]$ network provides a more accurate prediction during nearly the entire flight. Aside from a brief point at which the vehicle incorrectly predicts an obstacle, shown in Fig.~\ref{fig:obs-avoid-warehouse-flight-results}D, the network confidently predicts free space and avoids obstacles, yet trusting the $\mathcal{N}[\text{Simulated}]$ network would cause the vehicle to collide.

\section{Related Work}
\label{sec:rel-work}

There have been some recent efforts to train deep learning systems for real-world tasks using only simulated data. In \emph{Driving in the Matrix}~\cite{Johnson-Roberson:2017aa}, the high-budget game \emph{Grand Theft Auto V} was used to generate data for training an object detection system, since the simulated images are already rather realistic. Though using video games for image generation is an attractive approach, games are limited in versatility, making it difficult to generate images of environments or objects not present in the game world.

Some recent papers~\cite{domain-randomization,sadeghi2016cadrl} have attempted to directly encourage $p(l | i; D_s) \approx p(l | i; D_r)$, despite large differences in appearance between simulated and real images. By introducing large random variations in the lighting, color and texture of the simulated environments, they encourage the networks to learn color-invariant features like large-scale shapes or structure. However, the use cases of this approach are limited to tasks which do not rely upon texture and lighting, making it ineffective for tasks like semantic segmentation and is not applicable in general.

Finally, our work relies on recent progress in the domain of image translation~\cite{pix2pix2016} and style transfer~\cite{luan2017deep}. There have been some promising results in this space, yet most such approaches use pairs of similar or corresponding images, e.g.\ two images taken from the same vantage point in both the real and simulated world.
There are methods capable of relating \emph{unpaired} sets of images. One such result is~\cite{DBLP:journals/corr/ShrivastavaPTSW16}, in which the authors use a modified generative adversarial network to learn, among other things, a mapping from synthetic eyes to real eyes, and then use the resulting data for training simple tasks. Yet the construction of their \emph{refinement} network assumes low image variety and a close correspondence between the simulated and real-world images, making it difficult to use their approach to \emph{refine} complex scenes containing simulated objects whose shape, color or texture do not closely reflect the real world. The CycleGAN approach~\cite{CycleGAN2017} to unpaired image translation is designed with such scenarios in mind and is therefore an appropriate candidate for our work.

\section{Conclusions \& Future Work}
\label{sec:conclusion}

We have introduced \genesis{}, a procedure for generating realistic labeled synthetic images that can be used to train machine learning systems to perform real-world tasks. Using our mapping function $G_s$, trained using only unlabeled simulated and real-world images, can save the user days or weeks of time required to hand-label image data. We evaluated the quality of our \genesis{} data by using it to train neural networks for two different tasks: semantic segmentation and reactive obstacle avoidance. For both tasks we use for testing, we show that networks trained using our converted data are capable of outperforming those trained on raw simulated data alone. In the case of reactive obstacle avoidance, we additionally compare to networks trained on real-world data and show that our approach allows us to match and even exceed their performance. This is particularly true in our warehouse environment, in which we trust the network trained on our converted data to autonomously guide a quadcopter, thereby demonstrating that our data is of high enough quality to train neural algorithms to perform mission-critical tasks.

Evaluating performance on labeled real-world image data remains the best way to tune network parameters to maximize performance at test time, and doing so allowed us to improve performance for obstacle avoidance. Labeling a small amount of real-world data is a minor concession compared to the performance gains over using raw simulated data alone. Improving performance on real-world tasks in the absence of \emph{any} labeled real-world images during parameter tuning remains a future research goal. Finally, we have shown the effectiveness of our technique when we have a simulated model of the environment of interest, yet it remains a topic for future study to evaluate how well our approach to data generation will generalize to more generic environments.

\bibliographystyle{IEEEtran}
\bibliography{main.bib}
\end{document}